# Testing Hypotheses from the Social Approval Theory of Online Hate: An Analysis of 110 Million Messages from Parler

David M. Markowitz[1] & Samuel Hardman Taylor[2]

**Affiliations**

[1] Department of Communication, Michigan State University, East Lansing, MI 48824

[2] Department of Communication, University of Illinois Chicago, Chicago, IL 60607

## Abstract

We examined how social approval motivates online hate via the social approval theory, which argues social approval signals on hate messages predict more hate and toxicity. Using 110 million messages from Parler (2018-2021), we observed that upvotes on hate speech *posts* (e.g., messages broadcast on users' profiles to followers' feeds) predicted a lower probability of hate in the next immediate *post* but more in the following week's *post.* In other analyses, upvotes on *comments* (e.g., messages embedded within conversational threads) negatively predicted hate speech in the next *comment* but positively predicted toxicity during the next week and month. Between-person effects revealed a similar pattern of negative immediate short-term effects but positive longer-term effects for hate speech. For *comments*, social disapproval (downvotes) moderated these relationships, making them less positive at weekly through quarterly time-intervals and more strongly negative at six-months. Social approval predicts online hate in time- and message-dependent ways.

**Testing Hypotheses from the Social Approval Theory of Online Hate:**

**An Analysis of 110 Million Messages from Parler**

Online hate is vile and widespread on social media (United States Government Accountability Office, 2024; Walther, 2022). A recent report from the UCLA Bedari Kindness Institute, for example, found that approximately 80% of students encountered some form of hate speech on social media in the last month (Christie et al., 2024). Online hate is defined by messages that "express (or seeks to promote, or has the capacity to increase) hatred against a person or group of people" (Saleem et al., 2017, p. 1) specifically because of the target's identity (e.g., religion, gender, race, or ethnicity). It is perhaps unsurprising, then, to observe that over 70% of respondents from the same UCLA study encountered hate speech related to their gender and ethnicity. Together, many people report seeing hate speech on social media, and there is little evidence of its deceleration (Hickey et al., 2025; Madhu, 2025).

Because of the pervasiveness of online hate, research on this topic spans disparate fields and different theoretical perspectives. Indeed, much social scientific research has focused on the antecedents and consequences of online hate (Chen & Dang, 2023; Kazerooni et al., 2018; Sakki & Castrén, 2022), including motivations underlying why people engage in it (for a review, see Walther, 2022). Understanding the motivations and gratifications for participating in online hate is crucial for making sense of this phenomenon and providing insights into potential interventions to reduce aspects of social media that are considered toxic (Anjum & Katarya, 2024; Pradel et al., 2024). However, the motivations for participating in online hate are complex. The intent to harm another person or group may initially motivate online hate, and various individual differences (e.g., gender, psychopathy) (Bührer et al., 2024) can predict hate speech production as well.

Recent communication theory has offered an interpersonal explanation for the motivation, spread, and promotion of online hate on social media. The social approval theory of online hate (Walther, 2024) states that online hate is motivated by the goal of social approval, praise, and affirmation from others, rather than by motivations proposed in other work: the desire to inflict harm on or punish the victim (Paz et al., 2020; Ruscher, 2024). The theory's novel contention challenges conventional wisdom that animosity and maladaptive traits propagate online hate, and it warrants empirical testing and refinement above and beyond extant studies (Jiang et al., 2024; Rosen & Walther, 2025; Walther et al., 2025; Yang et al., 2025). The current work tests theoretical explanations of online hate motivations using a large collection of social media messages from an alternative social media platform with very limited moderation of online hate: Parler. Parler is considered an "alt tech" platform that facilitates extreme online communities (Dehghan & Nagappa, 2022; Freelon et al., 2020), and is similar to other alt tech platforms like Gab, Truth Social, and Rumble. It was considered a premier online space for "uncancelable free speech" from 2018 to 2021 (Cameron, 2023). For years after 2021, the platform was defunct because it was removed from Amazon Web Services and app stores due to insufficient content moderation associated with the January 6th insurrection on the US Capitol. After several relaunches, Parler has not regained the same user base nor cultural significance. Our paper used data during a time when the site was operational and culturally significant in America while also having minimal content moderation of hate speech (2018-2021).

There are several aims of the current work. First, we seek to provide empirical evidence that examines key tenets of the social approval theory of online hate and adds texture to its propositions and predictions. The theory argues that signals of social approval on online hate promote future online hate propagation. We examine the duration of this link at five discrete time periods: the next message, next week's messages, next month's messages, messages over the next three months, and messages over the next six months. We therefore examine how the theory and its premises operate longitudinally (see also Yang et al., 2025). Second, prior work suggests online hate proliferates in different types of social media messages (Kolluri et al., 2025; Rosen & Walther, 2025). We specifically examine how social approval links to online hate in *posts*, or messages broadcast on users' profiles to followers' feeds, and *comments*, or messages embedded within conversational threads. Third, the social approval theory suggests signals of social disapproval, such as downvotes or disapproving messages, have no direct effect on future online hate without additional praise from admirers. This contention raises an opportunity to determine if signals of social approval and disapproval jointly impact hate speech production. Together, we test the social approval theory by investigating the impact of time, message type, and social disapproval effects on hate speech propagation.

**Motivations for Online Hate**

Why do people communicate hate speech on social media? This interest in hate speech motivations has historically been addressed from two main perspectives (see Walther, 2022). The first perspective uses enduring psychological characteristics of the online hate perpetrator (e.g., individual differences) to form a profile of who communicates online hate and why they perform such acts. For example, prior work has linked sadistic personality traits and other aversive personality traits (e.g., The Dark Triad; Furnham et al., 2013) to online trolling (Buckels et al., 2019) and online hate (Sorokowski et al., 2020). These investigations are often labeled "bad actor" studies as they suggest the motivations for online hate propagation come from within the individual, with bad people performing hateful actions (Blackwell et al., 2018). The second perspective relates to social identity (e.g., Tajfel & Turner, 1979). That is, people tend to form connections with those who share similar characteristics and develop favoritism toward their in-group while displaying prejudice or hostility toward out-groups (Tajfel & Turner, 2004). From this perspective, hate speech is a mechanism for reinforcing group boundaries and asserting in-group superiority. Early research has demonstrated that social identity threats can trigger defensive intergroup hostility (Branscombe et al., 1999), and that online environments may amplify these dynamics where in-group norms favoring derogation of out-groups become intensified (Chatzakou et al., 2017; Williams & Burnap, 2016). Unlike the "bad actor" perspective, the social identity perspective suggests ordinary individuals may engage in hate when group-based motivations are activated, independent of their individual psychological traits.

Against this backdrop, a new and novel theoretical perspective explicating the motivations for online hate originates from Walther's (2024) social approval theory of online hate. The theory suggests social approval — "praise, affirmation, congratulations, reciprocation, and the like" (Walther, 2024, p. 6) — from likeminded audience members is a primary motivator for hate speech. Hate posters craft messages to attract positive responses from others and the theory focuses on the accrual of social approval to one's own hate messages. The social approval theory therefore envisions online hate as an audience-aware process for the perpetrator, focusing on the recruitment of others' praise and less about harming or punishing a target.

The social approval theory of online hate has six propositions and hypotheses, several of which are core to our interests. The first proposition is based on operant conditioning and the idea that signals of social approval facilitate a positive reinforcement cycle. Likes, upvotes, or other forms of engagement are social approval signals of "social acceptance" (Rosenthal-von der Pütten et al., 2019, p. 76), and thus, accruing them on one's own hate messages encourages that person to engage in more hate. Thus, the foundational tenet of this theory predicts that more signals of social approval (e.g., upvotes) on one's hate messages predicts more subsequent hate messages. An extension of this hypothesis suggests signals of social approval predict more extreme subsequent hate messages, as the extremity may engender additional praise and attention. Extreme hate messages are conceptually defined as messages with especially vile and negative sentiment (Cheng et al., 2017), and the social approval theory of online hate's prediction is based on studies indicating social media users communicate more toxic messages (Shmargad et al., 2022) and more morally outrageous *posts* (Brady et al., 2021) upon receiving more social approval signals in prior *posts*. The theory proposes as social approval increases, hate speech will become more extreme and toxic.

Initial empirical tests of the social approval theory of online hate have commenced. Yang and colleagues (2025) selected a random sample of 1,000 hateful and toxic *posts* (plus, over 1 million subsequent *posts*) from Gab, an alternative microblogging and social networking site (St. Aubin & Stocking, 2023), to examine how toxicity increases as a result of accruing likes and "receiving affirmation replies to one's hate posting" (p. 11). Inconsistent with the social approval theory of online hate, there was no significant link between the number of likes received on hate *posts* and the toxicity of one's subsequent *post*, nor was there a significant link between likes on hate *posts* and the toxicity over the next three months. Rosen and Walther (2025) also evaluated replies received on anti-Muslim and anti-Jewish messages from the social media platform, X. Replies convergent with an original hate *post* — that is, messages containing "a similar way of speaking with other members of the same group" (Rosen & Dale, 2024, p. 3140) — led to more hate speech in subsequent *posts*. Signals of social approval, such as likes, also precipitated hate posting. A related investigation evaluated over 1.2 million Tweets from members of the US government and observed that uncivil Tweets (Frimer et al., 2023), defined as those with impolite language that most individuals would find disrespectful (Mutz & Reeves, 2005), received more signals of social approval ("likes") than civil ones. Further analyses of these social approval markers indicated that "politicians learned from their positive experiences with uncivil tweeting and thus doubled down" (Frimer et al., 2023, p. 265). Together, there is some supporting evidence for the link between social approval and online hate across various platforms, although ambiguities remain and we attempt to examine them in the present work.

Clarifications regarding the function of social approval in online hate rest on (1) testing various time scales of subsequent online hate propagation, (2) differentiating social media message types that contain hate, and (3) evaluating the joint impact of social approval and disapproval on future hate production. We mainly focused on the idea that social approval predicts more frequent hate and more toxicity because such claims are vital for the utility of the theory. We also examined the joint impact of social approval (e.g., upvotes) and social disapproval effects (e.g., downvotes) in Parler *comments*.

**Research Questions**

Our study aimed to determine to what extent signals of social approval drive online hate propagation as the social approval theory predicts. If social approval reinforces hate speech production, this could indicate that platform design features (e.g., likes) may, counterintuitively, amplify hateful content and be mechanisms targeted for hate speech mitigation. However, our dataset — millions of Parler *posts* and *comments* — facilitated an exploration into how social approval effects are linked to online hate as a function of time, message type, and the associated design features of different message types.

We first test the temporal nature of the relationship between social approval signals and hate speech. The examination of multiple time scales is grounded in operant conditioning, which argues that the strength and persistence of a behavior often depend on its reinforcement pattern (Skinner, 1974; Thorndike, 1913, 1932). As a review by Burgoon et al. (1981) suggests, operant conditioning (i.e., behavior shaped by reinforcement) differs from classical conditioning (i.e., reflexive responses elicited automatically by stimuli) by introducing "a consequence, which impacts on the probability of the preceding response's occurring again in the presence of the same stimulus" (p. 167). Crucially, reinforcement effects do not operate uniformly over time (e.g., Ferster & Skinner, 1957; Skinner, 1974). When applied to online hate, the accumulation of social approval signals (e.g., upvotes) may require time to manifest behavioral effects (e.g., hate speech propagation). Shorter temporal windows (the next message, weekly) may capture immediate reinforcement dynamics where recently received approval directly influences subsequent posting behavior. However, longer temporal windows (monthly, quarterly, semi-annually) allow for the gradual accumulation of social approval that may be necessary to produce stable behavioral effects.

Evaluating different time thresholds for the accrual of social approval is consistent with the operant conditioning logic guiding the social approval theory of online hate. Operant conditioning emphasizes that reinforcement alters the probability of a behavior recurring. However, the drive-reduction model (Hull, 1943, 1952) suggests habit strength, drive, and incentive motivation must exceed a certain threshold before behavior is expressed. This idea is directly relevant to online hate propagation. Users may not encounter all accumulated social approval signals before posting their next message, meaning that immediate time periods may fail to capture reinforcement effects because incentive motivation is low. According to this theoretical position, only when users return to the platform over weeks or months would they encounter sufficiently accumulated approval signals, which may then exceed the motivational threshold necessary to strengthen hate speech behavior. Crucially, different temporal windows may capture fundamentally different psychological processes. Immediate effects may reflect reinforcement that users have already encountered, while longer-term effects may reflect the gradual accumulation and encounter with approval signals over repeated platform visits. Recent research on online hate motivations supports the necessity of thinking how social approval mechanisms change with time. For example, Yang et al. (2025) found that aggregated social approval signals predicted less toxicity in the short term, but were not associated with toxicity over the next three months. The present analysis tests social approval effects at five distinct temporal intervals to identify the time scale(s) at which accumulation produces observable behavioral change.

RQ1: What is the relationship between social approval and hate speech production at varying time intervals for *posts* and *comments*?

Social approval may take time as an accrual process in online hate, and this process may not be uniform across all types of social media messages in which people communicate online hate. Thus, we examine how different message types on Parler (e.g., *posts*, *comments*) might facilitate differential social approval effects. While *posts* may differ from *comments* on several characteristics, perhaps the most consequential of them concerns the idea that *posts* and *comments* on Parler create reinforcement environments that may produce different social (dis)approval effects on hate speech propagation. Previous research has identified that social approval in *comments* (or replies) can contribute to the propagation of online hate (Rosen & Walther, 2025) and therefore, we address an opportunity to build on this work by determining where on social media the effects of social approval on hate speech are most potent. On Parler, the unidirectional feedback on *posts* via upvotes provides largely positive reinforcement signals.

For *comments*, however, reinforcement is bidirectional with upvotes and downvotes as possible forms of engagement. This asymmetry creates different structures for learning what incentivizes online hate propagation (Hull, 1943, 1952; Walther, 2024). For *posts*, hate posters experience only a reward (e.g., upvotes) or the absence of a reward; for *comments*, downvotes introduce ambiguity into the reinforcement signal. Hate commenters must parse mixed feedback and determine how the quantity of approval and disapproval constitutes sufficient reward (or if it does at all). These differences may create distinct reinforcement environments. Given these distinct reinforcement mechanisms, we analyze social approval effects separately for *posts* and *comments* to determine whether the relationship between social approval signals and subsequent hate speech production varies across these functionally different communication and message-related settings.

RQ2: To what degree are signals of social approval on hate speech *posts* and *comments* linked to subsequent hate speech production?

We next address the idea that hate speech messages will become more extreme as social approval increases. We evaluated this through a continuous measure of toxicity, defined as a "rude, disrespectful, or unreasonable comment that is likely to make people leave a discussion" consisting often of personal attacks (Fan et al., 2024; Perspective API, 2023; Wulczyn et al., 2017). Based on our interests in the factors altering social approval mechanisms of hate speech propagation on Parler, we examine toxicity on *posts* and *comments* as well as examining the temporality of these effects:

RQ3: What is the relationship between signals of social approval and subsequent toxicity for *posts* and *comments*?

RQ4: What is the relationship between signals of social approval and subsequent toxicity at varying time intervals for *posts* and *comments*?

When only positive feedback is possible online (e.g., *posts* only allow upvotes and not downvotes), users generally receive signals that their hate speech is approved by the community, which may reinforce future hate speech production. When both positive *and* negative feedback are possible (e.g., *comments* allow upvotes and downvotes), users may receive mixed signals that can complicate the reinforcement process. A question, therefore, surfaces: To what degree is the effect of social approval (on hate speech production) contingent on disapproval mechanisms as well? The social approval theory also argues that social disapproval has no direct relationship with online hate unless onlookers support such disapproval with their own approval mechanisms (e.g., upvotes). If so, social disapproval is expected to be associated with more hate messages and more toxic speech. To date, the most direct test of social approval and disapproval working in conjunction suggests the opposite of their theory's prediction: receiving approval and disapproval predicted less subsequent online toxicity, rather than more toxicity (Yang et al., 2025). We test this contention by directly investigating the joint impact of social approval and disapproval on their connection with online hate and toxicity, longitudinally.

RQ5: What is the joint impact of social approval and social disapproval in predicting online hate and toxicity at varying time intervals for *comments*?

# Method

## Data Collection and Preprocessing

To test the social approval theory's assertions, we analyzed a historical collection of Parler messages (Aliapoulios et al., 2021). Recall, Parler was an "alt tech" platform that marketed itself as an uncensored social networking site, specifically for conversations from alternative viewpoints often removed from mainstream social media like Facebook or Instagram (Stevenson et al., 2024). At the time the data were collected (2018-2021), Parler was the most popular alternative social networking site, used primarily among conservatives in the US (Blazina & Stocking, 2022; Harwell & Lerman, 2020). Parler was often positioned as a platform for free speech by users and supporters (Stocking et al., 2022), and approximately one-third of its users would claim to see at least some unfriendly discussions on the site (Blazina & Stocking, 2022). The platform became associated with conspiracy theories, election misinformation, and extremist organizing, culminating in its temporary shutdown following the January 6, 2021 attack on the US Capitol. Since 2021, Parler has relaunched numerous times under new ownership with stricter content policies. Our data, containing millions of Parler messages from millions of users between August 2018 and January 2021 (Aliapoulios et al., 2021), represent the original, largely unmoderated version of the platform, which provides a valuable setting for testing social approval theory dynamics of online hate (but may contain different moderation tactics and user bases than other platforms). The dataset has been used to evaluate hate-speech-adjacent topics like the spread of racist content (Kolluri et al., 2025), with evidence suggesting one's engagement with racist content increases the probability of a person posting racist content.

The Parler dataset contains two types of social media messages that can receive social approval or social disapproval. First, broadcast and undirected *posts* — similar to status updates or Tweets that can receive only upvotes, reposts[1], and directed comments — are henceforth referred to as *posts*. There were no downvotes for *posts* on Parler. Second, directed *comments* threaded onto *posts* are henceforth referred to as *comments*. For *comments* only, the message receives a tally for the number of upvotes received and another separate tally for the number of downvotes received.

[1] We did not evaluate reposts because it was unclear if they met the formal definition of signals of social approval. People may repost a message for many reasons (Metaxas et al., 2015), independent of their approval of the message.

**Hate Speech and Toxicity Identification**

Our study evaluated how social approval links to two different constructs: (1) hate speech and (2) toxicity. We used a dictionary of terms curated by the Weaponized Word to identify hate speech in Parler messages; any message containing at least one of these terms was considered a hate speech message. The Weaponized Word and its data, which have been used by organizations like The Office of the United Nations High Commissioner for Human Rights, maintains a dictionary of 1,598 words that "disparage a specific targeted group or identity" (Weaponized Word, 2025b), which is consistent with conceptualizations of online hate (Hietanen & Eddebo, 2023; Saleem et al., 2017). Given their harmful nature, these words are not repeated here, but they include racial epithets, religious slurs, and other terms that clearly meet the definition of hate speech. This collection of words represented our hate speech dictionary.

Toxicity was measured via the Detoxify classifier (Hanu & Unitary, 2020). This Bidirectional Encoder Representations from Transformers (BERT) architecture (specifically, *bert-base-uncased*) (Devlin et al., 2019) was fine-tuned on the Jigsaw Toxic Comment Classification dataset and is a natural language processing model trained on data labeled by human annotators who rated messages for toxicity (Fan et al., 2024; Perspective API, 2023; Wulczyn et al., 2017). The model, which contains over 110 million trainable parameters, estimates the probability that a message contains toxic speech. The resulting score is a continuous measure from 0 (no toxicity) to 1.00 (extreme toxicity) with higher scores indicating greater toxicity. All messages from the Parler dataset were run through the Detoxify classifier. To demonstrate the toxicity classifier in practice, we offer a selection of deidentified cases: "[Name] is a lyin POS!" (toxicity score = 0.839899); "Catch them and throw them all in prison" (toxicity score = 0.511676); "Time to boycott Amazon" (toxicity score = 0.100338).

### *Validity Checks*

The Weaponized Word's collection of hate speech terms is human-generated. As the Weaponized Word suggests, their lexicons are "curated by staff and volunteers with the assistance of regional partner organizations around the world" (Weaponized Word, 2025a). Given the human-generated nature of this collection, it was unnecessary to revalidate the terms as hate speech, especially given the potentially harmful and unintended consequences that reading messages with such terms might have on individuals (AlEmadi & Zaghouani, 2024). Still, we performed validity checks on the Weaponized Word dictionary by randomly sampling 1,000 cases from the Parler dataset, half of which were identified as hate speech (e.g., terms from the Weaponized Word dictionary > 0) and the other half that were identified as not hate speech (e.g., terms from the Weaponized Word dictionary = 0). These cases were submitted to annotation by a large language model (LLM), *gpt-4o*, via the OpenAI API and the prompt solicited a scalar measure of online hate as indicated by prior work (Walther et al., 2025). Given that LLMs are "trained on collective data that encapsulate the contributions of countless individuals" and "prompting an LLM with a question is like a distilled form of crowdsourcing," we aimed to identify if messages coded as hate speech from the Weaponized Word would be labeled as hate by an entity that can approximate "collective intelligence" on perceptions of hate speech (Burton et al., 2024, p. 1645). LLM performance on such hate speech identification tasks often meets or exceeds human abilities (Fasching & Lelkes, 2025; Huang et al., 2023; Shi et al., 2024). Together, from prior work (Walther et al., 2025), the prompt stated:

> *Annotate the following text for level of hate using integers from 1 to 6, based on the following definition: If there are no hateful or derogatory words and the message has a neutral or positive tone, annotate it as 1. If there are mildly derogatory words with a* *slightly negative tone against certain groups or individuals, annotate it as 2. If there is a moderately hostile or derogatory language against certain groups or individuals, with a quite negative tone, and the purpose is to belittle certain groups or individuals or the text manifests prejudice, annotate it as 3. If there is strongly hostile or derogatory language against certain groups or individuals, and the purpose is to belittle certain groups or individuals, but there is no advocacy for actual harm, annotate it as 4. If there is extremely hostile, derogatory, or dehumanizing language against certain groups or individuals with a clear purpose to disparage or attack the character or degrade such groups or individuals, and it advocates greater malevolence or antipathy, but there is no advocacy for actual harm, annotate it as 5. If there is extremely hostile, derogatory, or dehumanizing language with a clear purpose to attack or degrade such groups or individuals, which also recommends physical aggression and violence, even extinction against such groups, annotate it as 6. Return the result in the form of the numeric rating.*

Using our binary classification of hate speech that is represented in the analytic models described below (e.g., a message either contained a hate speech term from the Weaponized Word or it did not), messages identified as hate speech by containing terms from the Weaponized Word ($M$ = 2.39, $SD$ = 1.26) were also identified by *gpt-4o* as containing more hateful speech than messages not identified as hate speech ($M$ = 1.48, $SD$ = 0.90) ($t$ = 13.04, $p$ < .001, Cohen's $d$ = 0.836). Further, this binary hate speech measure and a second Detoxify measure that assesses the degree to which a message is an *identity attack*[2] (e.g., "negative or hateful comments targeting someone because of their identity"; Korre et al., 2025; Perspective API, 2023) were significantly related as well ($t$ = 14.82, $p$ < .001, Cohen's $d$ = 0.937). Therefore, the LLM-coded data and human-annotated identity attack classifier demonstrate the Weaponized Word dictionary is operating as intended.

[2] The identity attack measure was scaled from 0.0 (no identity attack) to 1.0 (extreme identity attack).

**Analytic Plan**

According to this paper's preregistration (https://aspredicted.org/ghqp-zxcq.pdf), messages containing at least one word from the Weaponized Word list were coded as hate speech. From the full sample of approximately 180 million messages, we excluded cases where no words were present ($n$ = 67,459,674 messages). The final dataset contained 3.7 million hate speech messages, or 3.4% of the final collection ($N$ = 111,393,499 messages).[3] Among the 3.7 million hate speech messages, 22.2% were *posts* and 77.8% were *comments*. This final sample originated from 4,000,322 unique creators[4] and contained nearly 2 billion words. See Table 1 for descriptives and Supplementary Figure S1 for distributions of hate speech and toxic messages.

***Analysis 1***

We conducted two main analyses. First, we used a multilevel mixed effects logistic regression (Bates et al., 2015; Kuznetsova et al., 2020) to test our research questions on messages coded as hate speech and the propagation of hate speech from message-to-message. We identified every message coded as hate speech by the Weaponized Word and paired it with the same creator's next immediate message within the same message type (posts were paired only with posts, and comments were paired only with comments).[5] Hate speech messages that were a creator's final message were excluded because no subsequent message was available.

[3] To contextualize this number, we evaluated hate speech rates in other studies. In one investigation, 2.5% of Tweets about China during the COVID-19 pandemic contained hate speech (Xu & Weiss, 2022), with other estimates of online hate being between 4% and 30% of online messages depending on the country and the online platform (Hine et al., 2017). Therefore, the rate of online hate in the present paper falls within range of other studies.

[4] Creators are people who published *posts* or *comments* on Parler.

[5] A creator's next message could be in the same day or weeks later. In our data, less than 1% of pairs were separated by more than 30 days.

Therefore, in this analysis, we assessed: To what degree does the amount of social approval (i.e., upvotes) received on a person's prior hate speech message ($t - 1$) predict the probability of their next message being hate speech ($t$)?

Predictors were lagged by one message, and then group-mean centered to capture within-person and between-person associations. Within-person effects indicate whether a message receiving more or less upvotes than is typical for that creator is associated with an increased or decreased probability that the creator's next message contains hate speech; coefficients represent deviations from the Parler creator's mean. Between-person effects indicate whether creators who receive more upvotes on their hate speech posts, on average, differ in their average probability of communicating hate speech; coefficients are aggregated associations across their messages in the analytic sample.

Our dependent variable was a binary hate speech presence (coded as 1) versus absence (coded as 0) measure, where presence is any message containing at least one term from the Weaponized Word dictionary. Predictors were standardized (z-scored), and we included a random intercept for the creator to control for non-independence.[6] This analysis was performed on *posts* and *comments* separately.

***Analysis 2***

The second analysis also used the full dataset of 110 million messages to evaluate the time-bounded nature of these effects. The dependent variable in this analysis was the percentage of one's *messages within a specified time period* that contained any hate speech, natural log-transformed after adding a constant, *ln*(hate percentage + 1) out of skewness concerns. For example, if a person produced 10 *comments* on Parler in one week, and four of the *comments* had hate speech terms in them from the Weaponized Word, a person's raw score for that week would be 40% (e.g., 40% of their *comments* from the week contained hate speech). Therefore, Analysis 2 answers the question: To what degree does the amount of social approval (i.e., upvotes) received on hate speech messages in a prior time period ($t-1$) predict the percentage of one's messages containing hate speech in the following period ($t$)?

[6] We attempted to control for conversational thread with a random intercept (for *comments*, only), but the models did not achieve sufficient fit with this random effect included. It was therefore dropped from these specifications.

In multilevel mixed effect regression models, predictors were time lagged by one of four time points depending on the model (i.e., weekly, monthly, three-month or quarterly, six-month or semi-annually). We chose these time intervals because it tests short-term and long-term consequences, and it is common in hate speech research to aggregate messages monthly and longitudinally (Garland et al., 2022; Hickey et al., 2025; Mathew et al., 2020; Siegel et al., 2018). Thus, four models were constructed to evaluate how the total sum of upvotes on hate messages in the last week, month, three months, or six months predicts the percentage of their messages containing hate speech in the following week, month, three months, or six months. Predictors were group-mean centered to capture within-person and between-person associations. Predictors were also standardized in all models and multilevel models contained a random intercept for the creator to control for non-independence.

Consistent with Analysis 1, separate models for *posts* and *comments* were constructed. We did not control for conversation because of the aggregation of such time-related analyses. Time lags are consecutive periods of activity. Users who were inactive during intervening periods are included, with their "next period" being whenever they next *posted* or *commented*. An overview of our work — including predictions, research questions, and methodological approaches — is represented in Figure 1.

## Results

**Social Approval Effects on Hate Speech**

Inconsistent with the social approval theory of online hate (Walther, 2024), the number of upvotes a person received on a hate speech *post* was negatively associated with the probability of their next *post* containing hate ($B$ = -0.039, $SE$ = 0.01, $z$ = -4.16, $p$ < .001, Odds Ratio (OR) = 0.96). Between-person effects revealed that creators who received more upvotes on their hate *posts*, on average, were less likely to produce hate speech in their next *post* compared to creators who received fewer upvotes ($B$ = -0.026, $SE$ = 0.01, $z$ = -2.47, $p$ = .014, OR = 0.97). For *comments*, the number of upvotes received on a hate speech *comment* was negatively associated with the probability that one's next *comment* contained hate speech ($B$ = -0.004, $SE$ = 0.002, $z$ = -2.22, $p$ = .026, OR = 1.00). Between-person effects revealed that creators who received more upvotes on their hate *comments*, on average, were less likely to produce hate speech in their next *comment* compared to creators who received fewer upvotes ($B$ = -0.02, $SE$ = 0.003, $z$ = -5.97, $p$ < .001, OR = 0.98). For a visual description of the top 20 hate speech communicators disaggregated by message type (*posts* vs. *comments*) see Supplementary Figure S2.

Next, we evaluated the relationship between upvotes and hate speech at different time intervals beyond the message-to-message level. *Posts* and *comments* were disaggregated, and we observed inconsistent effects across these message types. The number of upvotes a person received on their hate speech *posts* in the past week was positively associated with the amount of hate speech in their subsequent week's *posts* ($p$ < .001; see the top panel of Supplementary Table S1). However, this pattern was not statistically significant at the monthly, quarterly, and semi-annual time intervals. Between-person effects for upvotes in the previous weekly, monthly, quarterly, and semi-annual time intervals were positively related to hate speech and statistically significant in all models. The bottom panel of Supplementary Table S1 reveals a positive relationship between upvotes and hate speech production for *comments* at the weekly time period, but a nonsignificant relationship between upvotes and hate speech production at the monthly time period ($p = .084$), and a negative and significant relationship at the quarterly ($p < .001$) and semi-annual time periods ($p < .001$); all between-person effects were positive and statistically significant ($p < .001$). Therefore, for *comments*, social approval is positively linked with hate speech production at the shorter weekly level, but it was negatively associated with hate speech production at longer time intervals.

**Social Approval Effects on Toxicity**

Addressing RQ3, the number of upvotes a person received on their prior hate *post*[7] was statistically unassociated with the probability of producing a toxic message in their next *post* ($B = -0.012$, $SE = 0.009$, $z = -1.30$, $p = .193$, OR = 0.99). Between-person effects revealed that creators who received more upvotes on their hate *posts*, on average, were less likely to produce toxic speech in their next *post* compared to creators who received less upvotes ($B = -0.125$, $SE = 0.02$, $z = -6.91$, $p < .001$, OR = 0.88). At the within-person level, the number of upvotes received on their prior hate *comment* was negatively associated with the probability of producing a toxic message in their next *comment* ($B = -0.021$, $SE = 0.002$, $z = -9.43$, $p < .001$, OR = 0.98). Between-person effects revealed a consistent negative average effect as well ($B = -0.08$, $SE = 0.003$, $z = -23.98$, $p < .001$, OR = 0.92).[8]

Addressing RQ4 (Supplementary Table S2), upvotes and toxicity were statistically unassociated for *posts* at both the within- and between-person levels across all four aggregated time intervals.[9] For *comments*, at the weekly and monthly time periods, there were positive associations between upvotes and toxicity at the within-person and between-person levels. At the quarterly and semi-annual time periods, there were no significant relationships between social approval and toxicity at the within-person level, but positive associations between social approval and toxicity at the between-person level.

[7] For the next-message analyses with toxicity, we dichotomized toxicity based on conventions of prior work: scores > 0.50 were deemed toxic; otherwise, messages were deemed non-toxic (Jigsaw, 2019). Therefore, like the message-to-message analysis with hate speech, we evaluated how upvotes on a prior hate speech message predicted the probability of one's subsequent message being toxic.

[8] Treating toxicity as continuous in the message-to-message analyses produced consistent results in terms of direction and statistical significance.

**Social Disapproval Effects on Hate Speech and Toxicity**

Addressing RQ5, we created lagged downvotes at the within and between-person levels, and created interaction effects with respective lagged upvotes at the within and between-person levels. A binary logistic regression mixed model was computed, controlling for creator as a random intercept, and this was calculated for *comments* only since downvotes only appear for this message type.

Predicting hate speech on the next *comment*, there was a statistically significant interaction effect for the within-person parameter ($B$ = 1.454e-04, $SE$ = 4.658e-05, $z$ = 3.12, $p$ = .002, OR = 1.00) and the between-person parameter ($B$ = -4.618e-03, $SE$ = 3.960e-04, $z$ = -11.66, $p < .001$, OR = 1.00). This evidence suggests the relationship between upvotes and hate speech *comments* depends on downvote levels, but this varies within the same person over time and between people. When predicting toxicity, there was a statistically significant interaction effect for the within-person parameter ($B$ = 2.504e-04, $SE$ = 4.307e-05, $z$ = 5.82, $p < .001$, OR = 1.00) and the between-person parameter ($B$ = -6.016e-03, $SE$ = 2.559e-04, $z$ = -23.51, $p < .001$, OR = 0.99). Within-person, the negative association between upvotes and next-*comment* hate speech and toxicity weakened when more downvotes accompanied the prior hate *comment*. Between-person, the negative association strengthened among creators who received more downvotes, on average. Please see Supplementary Table S3 for estimated marginal means.

[9] For the time-lagged toxicity models, the dependent variable was the mean Detoxify toxicity score across all a creator's messages in the period (of that message type), re-expressed as *ln*(mean toxicity + 1).

The evidence at other time intervals revealed a complex picture for hate speech *comments*. Supplementary Table S4 reveals statistically significant and negative interaction effects between upvotes and downvotes at the within-person level across weekly, monthly, quarterly, and semi-annual time periods for both hate speech and toxicity (see Supplementary Table S5 for estimated marginal means). These negative interactions indicate that, as downvotes on prior hate *comments* increased, the within-person association between upvotes and subsequent hate speech and toxicity became less positive. At the weekly, monthly, and quarterly intervals, this reflects a weakening of the positive reinforcement effect of upvotes; at the semi-annual interval, where the within-person upvote association was already negative for hate speech (null for toxicity), greater downvotes were associated with a stronger negative association (see Supplementary Table S5). The between-person interaction effects were also statistically significant and negative at every time period. Together, these results indicate the effect of upvotes is contingent on the level of downvotes. Note, these effect sizes are very small, yet statistically significant likely due to our large sample size.

## Discussion

Drawing on the social approval theory of online hate (Walther, 2024), the goal of this paper was to analyze social media data to examine the notion that social approval from audience members perpetuates online hate. Our analysis of 110 million Parler messages demonstrated how and when social approval facilitates more online hate and toxicity. Three important dimensions of this empirical investigation have tested and advanced our understanding of online hate. First, we observed that depending on how messages are aggregated, the link between social approval and online hate changes. Our results provide large-scale, longitudinal evidence that online hate propagation is conditioned on social approval *and* time. Second, the results, disaggregated by *posts* and *comments*, showed different social approval effects in connection with hate speech production. Within-person evidence for *posts* demonstrated a negative message-to-message relationship between social approval and hate speech production but a positive week-to-week relationship. Upvotes on hate *comments* negatively predicted hate in the next *comment*, positively predicted the rate of hate in the following week, showed no significant association at the monthly interval, and negatively predicted hate at the quarterly and semi-annual timescales. Third, we evaluated the joint impact of social approval and social disapproval to predict hate speech and toxicity in *comments*. In every model except the within-person next-*comment* models, higher downvotes shifted the upvote slope in the negative direction. We dissect these findings below.

**Time-Related Effects**

This paper suggests the social approval theory can operate through the reinforcement mechanisms as originally proposed, but there are boundaries and contingencies that texture the theory. Individuals did not appear to uniformly increase hate speech production in response to social approval for either *posts* or *comments* over time, which is consistent with other work that was performed using hate posters on another fringe social network with limited moderation (i.e., Gab) (Yang et al., 2025). Within-person social approval effects for *posts* were mixed in the short-term (e.g., negative in message-to-message, positive in the weekly time-lagged analyses), and there was a null relationship between social approval and online hate for the monthly through semi-annual time periods. Only one significant relationship between social approval and *post* toxicity was observed: the average number of upvotes on a *post* predicted less toxicity in the creator's next *post*. *Comments* demonstrated a slightly different temporality: social approval was negatively associated with hate speech propagation at the *comment*-to-*comment*, quarterly, and semi-annual levels, but positively associated at the weekly level, suggesting different short- and long-term effects of social approval on future hate speech.

What social or psychological characteristics might help to explain these divergent patterns in hate speech production over time? At the message-to-message level, there may not be enough time for social approval signals to be accrued before the next message is produced (e.g., upvotes accrue gradually, and users may compose more *posts* or *comments* before sufficient social approval associated with a prior message has accumulated; Rosen & Walther, 2025; Törnberg & Törnberg, 2024). By the following week, however, users returning to the platform may encounter the aggregated approval that is attached to their prior hate *posts* and *comments*, which may increase the number of hate *posts* and *comments* in the next week. Therefore, the time needed to accrue social approval on hate or toxic content appears to be a central process to social approval theory that warrants further conceptualization.

This interpretation is consistent with the idea of a motivational threshold (Hull, 1952) in operant conditioning (Ferster & Skinner, 1957; Skinner, 1974), where the accrual of social approval may take some time (e.g., a week) to cross a threshold sufficient to reinforce behavior (cf. Yang et al., 2025). In this sense, increases in the likelihood of subsequent hate speech may reflect exposure to cumulative social approval over the course of a week rather than a single instance of approval. However, the reinforcing impact of approval also appears to be short-lived. Within-person associations were attenuated in monthly windows and became significantly negative for *comments* and non-significant for *posts* in quarterly and semiannual windows. One possibility is that these longer windows capture the attenuation of participation on Parler (e.g., users who eventually leave a platform may do so after active, high social approval periods), though testing this account of disengagement is a task for future research. The between-person relationships for hate speech, by contrast, were positive at every time interval, except message-to-message. People who received more approval on their hate *posts* or *comments* overall engaged in more subsequent hate speech, regardless of the aggregated time frame, supporting the notion that individuals who receive more social approval on their online hate messages tend to produce more online hate. Future work should more directly test how the impact of social approval on future hate speech may depend on accumulation and repeated encounters.

Several other compelling findings of this work are represented by the inconsistencies in the within-person effects and between-person effects in terms of their direction and statistical significance. Although the between-person relationships were largely stable over time, time intervals mattered for the direction and significance of the within-person correlations of social approval. This temporal discrepancy existed for the production of hate messages and message toxicity. These differences highlight an important question for the social approval theory of online hate: Over what time interval(s) does the accrual of social approval need to happen in order to affect or link to subsequent hate speech? Our study adds empirical evidence to demonstrate the social approval theory's time-dependent contentions are worth taking seriously and examining across platforms with online hate. Furthermore, our results are consistent with prior work that shows the time window considered for the accrual of social approval can alter if it is associated with subsequent hate (Yang et al., 2025). The temporality of the effect may be less linear than initially anticipated, as it is possible for a variety of longitudinal change relationships to exist (Brinberg & Lydon-Staley, 2023). Future research should examine the longitudinal nature and shape of social approval and online hate.

**Social Approval and Hate Speech Links by Message Type**

Another notable pattern in our work is the link between social approval and the propagation of hate speech and toxicity, as a function of message type. Despite some clear differences in the time-aggregated models, the effect of social approval on hate speech was more consistent across *posts* and *comments* than the effect of social approval on toxicity across message types. For example, when predicting toxicity, all time-lagged links between social approval and toxicity for *posts* were non-significant, and links between social approval and toxicity were positive and statistically significant for *comments* at the within-person level over the weekly and monthly time periods (yet negative for the *comment*-to-*comment* analysis). It appears that social approval in *comments* can encourage more rude or disrespectful attacks in shorter time periods compared to longer time periods, whereas this dynamic does not exist systematically in *posts*.

The different patterns of social approval and subsequent online hate and toxicity signal the need to consider how the motivation of seeking social approval intersects with technological features and affordances (see Bazarova & Choi, 2014). Upvotes on *posts,* perhaps, represent a more diffuse form of social approval than upvotes on *comments* because of the ambiguity of social meaning ascribed to such one-click social approval (Ellison et al., 2020). Receiving upvotes on *comments* may more directly communicate social approval to online hate communicators and therefore has a different social or psychological meaning than *posts*. One reason is that *comments* are conversational turns in an ongoing exchange, whereas *posts* are broadcast to followers' feeds. Threading in a conversation may also supply what hate speech or toxicity requires, as such vile attacks presuppose an interlocutor that *comments* have, and broadcast posts generally do not. Future work would benefit from a deeper theoretical understanding about why social approval on *posts* versus *comments* relates differently to subsequent hate and toxicity (e.g., what does a *comment* communicate or symbolize that a *post* does not?). Together, beyond hate speech *comments* being much more common than hate speech *posts*, our present study demonstrates there are meaningful differences in hate speech propagation as a function of the social media hate message.

**Social Approval and Social Disapproval Effects**

The social approval theory of online hate suggests signals of social disapproval indirectly increase online hate when it prompts sympathetic others to signal approval. Because of the explicit downvotes on Parler *comments*, these data allowed us to measure the joint impact of social approval and social disapproval in online hate propagation. Our results converged on a consistent pattern where downvotes attenuated the presumed motivational value of upvotes. At the between-person level, the interaction between social approval and disapproval was negative in every model for both hate speech and toxicity. Within-person, the same attenuation appeared across all time-lagged intervals, though not in the message-to-message models, where the interaction was positive. That is, as downvotes on prior hate *comments* increased, the association between upvotes and subsequent hate and toxicity weakened or grew more negative at the semi-annual time period. These results are consistent with Yang and colleagues (2025), who observed that mixed responses (e.g., likes and dislikes) reduced online toxicity. The effect sizes were small, and while we are cautious with our interpretations, small effects still matter at scale and can have a nontrivial impact on a population (Matz et al., 2017).

These results are important for our understanding of online hate and the social approval theory because they suggest an interplay between social approval and disapproval. Estimated marginal means indicated, for the time-lagged models, the association between upvotes and subsequent hate speech became progressively less positive as downvotes increased (see Supplementary Table S5). Only in the semi-annual time-period did it become more strongly negative within-person, and in the message-to-message analyses, the effect became less negative within-person but more negative between-person (see Supplementary Table S3). In operant terms, downvotes did not consistently function as a suppressive punishment. At the between-person level, creators receiving more downvotes tended to produce more subsequent hate and toxicity, not less. Instead, downvotes operated jointly with social approval (e.g., as downvotes increased, the association between upvotes and subsequent hate weakened). This pattern qualifies the social approval theory's treatment of disapproval. That is, disapproval need not generate support from admirers to matter for hate speech production, but its influence appears to attenuate the motivational value of simultaneous approval. Practically, this suggests community downvoting may temper the upvote-hate association rather than inflame it as the social approval theory suggests.

**Limitations and Future Directions**

Our tests of social approval and online hate research should be interpreted within the limitations of our study design. First, the social approval theory of online hate does not specify modeling procedures nor ways to best "chunk" the data in terms of time. Therefore, our results are perhaps bounded by the ways that we modeled the data, and future studies should consider other modeling approaches at varying time intervals. Recall, we attempted to account for within-conversation dependencies with a random intercept for conversation (in *comments*, only), but these models failed to converge and were thus unstable. We did not include these models to avoid the possibility of spurious results, but future work should attempt to account for messages nested in conversational threads if modeling allows. Second, although we offer some support for different aspects of the theory, we only assessed a refined number of hypotheses from the social approval theory of online hate. More empirical work, with platform and hate context sensitivities (e.g., audience composition), is required to provide definitive support (or lack thereof) for the hypotheses proposed by the social approval theory of online hate.

We were also limited to only one form of social approval: upvotes. It is possible that other forms of social approval (e.g., likes, reposts, or supportive comments) may produce different effects, but this is an open empirical question. Upvotes also do not exist on some social media platforms (i.e., Instagram), and it is also unclear *when* upvotes occurred for a message (or when users saw them). Evaluating these dynamics in online hate research is critical to deeply understand it propagation. Relatedly, we measured hate speech via the Weaponized Word dictionary. We validated this dictionary against two other measures, including ratings from a large language model that represents an important methodological advance, but future research should consider how different types of hate speech measures produce varying effects. Our validity checks ensured that the Weaponized Word collection was operating as we would expect for hate speech identification, however.

Finally, future experimental research should complement our paper to establish causality, and identify social and psychological mechanisms to further explicate the results. Future research should also consider how individual differences (e.g., personality, ideology) might impact such results, how those who propagate online hate bring their hate to many platforms within different social approval effects, and how cultural and critical moments in history play a critical role in the facilitation of hate speech (e.g., how hate speech might be impacted by cultural moments like a US presidential election, political and social unrest, and so on). These ideas should be examined in future work for a more nuanced view of social approval and hate speech production. Our account of social approval accumulation is also interpretive. The temporal patterning of effects is consistent with the gradual accrual of approval signals, but we did not model the rate or shape of that accrual, and testing accumulation processes remains an important task for future research.

**Acknowledgements**

We thank the editor and reviewers for their thoughtful and careful attention to our manuscript throughout the review process.

**Table 1**

*Descriptive Statistics for the Parler Sample*

| Measure | *M* | *SD* | *Median* | Min | Max |
|---|---|---|---|---|---|
| Hate speech messages per creator | 0.93 | 19.72 | 0.00 | 0.00 | 25,127 |
| Messages with hate speech (%) | 1.55 | 7.76 | 0.00 | 0.00 | 100 |
| Upvotes per message | 4.95 | 157.28 | 1.00 | 0.00 | 240,000 |
| Downvotes per message (comments) | 0.09 | 1.46 | 0.00 | 0.00 | 962 |
| Toxicity | 0.14 | 0.29 | 0.00 | 0.00 | 1 |
| | *n* | | | | |
| Number of creators | 4,000,322 | | | | |
| Number of cases | 111,393,499 | | | | |

*Note*. Descriptive statistics pertain to the sample of 111 million messages. Toxicity scores were obtained using the Unitary AI Detoxify model. Downvotes are relevant only to *comments* in each analysis and sample. Some models across analyses might have slightly smaller counts due to missing data. Messages with hate speech refers to the average percentage of messages that contained hate speech for each person.

**Figure 1**

*Research Overview*

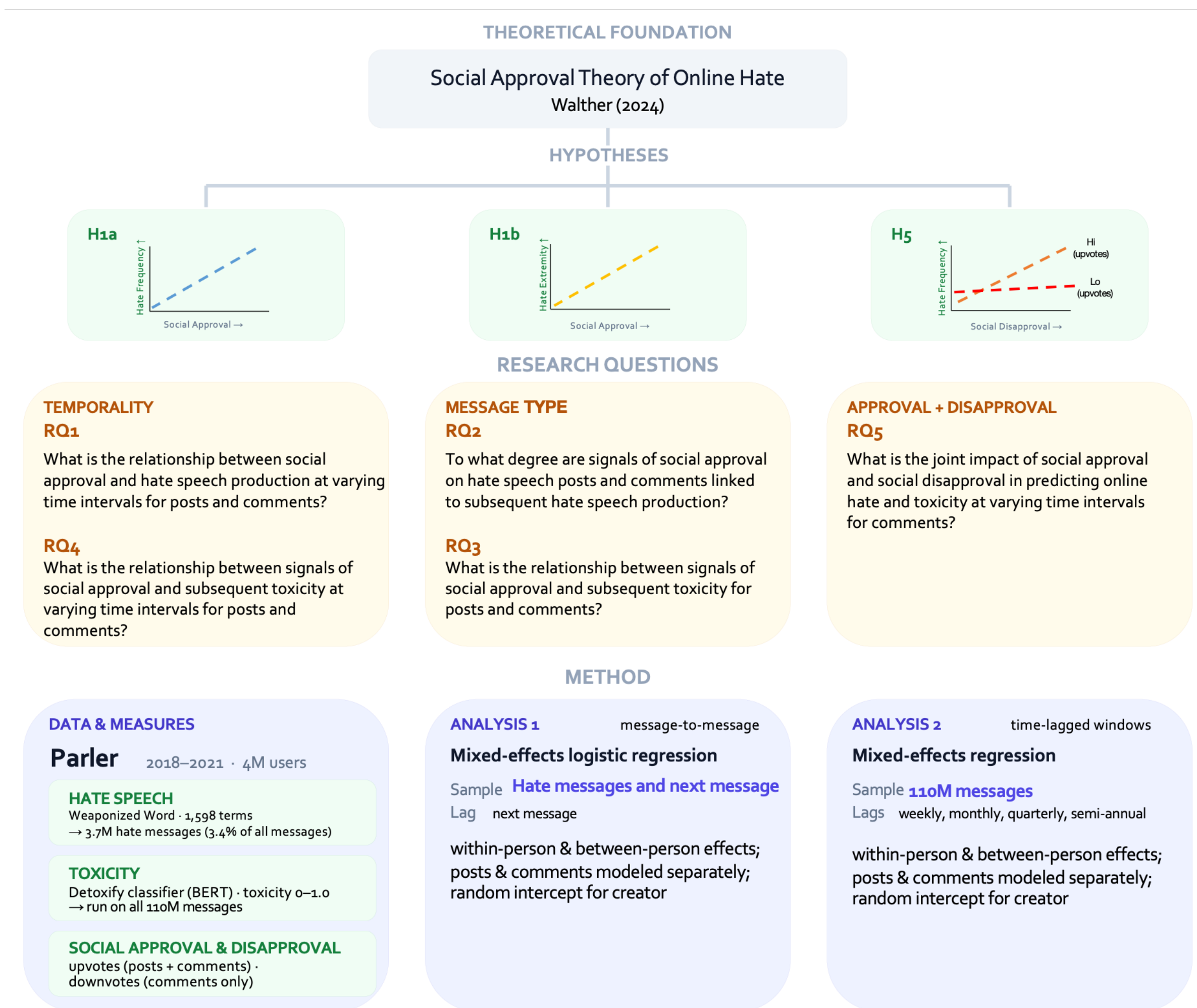

*Note*. This figure is a simplified distillation and overview of the present work. Labels for the hypotheses correspond to those represented in the social approval theory of online hate. We used artificial intelligence (Claude Sonnet 4.5) for inspiration regarding Figure 1. We then adjusted the figure and text to ensure accuracy and readability.